\documentclass[11pt]{article}
\usepackage[margin={2cm,2cm}]{geometry}
\usepackage{multicol}
\usepackage{cite}
\usepackage{amsmath}
\usepackage{float}
\floatstyle{boxed}

\usepackage{hyperref}
\usepackage{xcolor}
\hypersetup{
    colorlinks=true,
    linkcolor=blue,
    filecolor=magenta,
    urlcolor=cyan,
}

\usepackage{csquotes}

\usepackage{graphicx}
\graphicspath{ {./images/} }

\begin{document}

\title{Cross Fertilizing Empathy from Brain to Machine as a Value Alignment Strategy}
\author{Devin Gonier (dcg2151), Adrian Adduci. (faa2160) ,Cassidy LoCascio (cll2182) }
\date{2021-12-18}
\maketitle

\begin{multicols*}{2}
    \section{Abstract}

    AI Alignment research seeks to align human and AI goals to ensure independent actions by a machine are always ethical.
    This paper argues empathy is necessary for this task, despite being often neglected in favor of more deductive
    approaches. We offer an inside-out approach that grounds morality within the context of the brain as a basis for
    algorithmically understanding ethics and empathy. These arguments are justified via a survey of
    relevant literature. The paper concludes with a suggested experimental approach to future research and some initial
    experimental observations.

    \section{Introduction}

    Within the field of Artificial Intelligence (A.I.), the hope is that humanity can construct algorithms that behave in
    ways that represent human values. A.I.'s consistently impress people with monumental feats in strategy, modeling,
    prediction, and more. Still, even the best algorithms struggle with basic everyday tasks like humor. The relative
    underperformance of current technology is mainly because some of the most impressive human gifts occur seamlessly at
    subconscious levels.

    As A.I. abilities grow, so do their effects, which have the potential to either liberate the human race from many of
    our most significant forms of suffering or endanger our entire species. While the field of A.I. ethics is burgeoning,
    with many sounding the siren call, the more emotive elements of moral reasoning are often neglected. We believe a
    central element of human morality is empathy, and we aspire to find ways to manifest it in machines.

    Our primary argument is that moral decision-making requires empathy and is, therefore, a necessary component of
    building artificial moral intelligence. We propose a research paradigm that bases moral reasoning in neuroscience and
    attempts to build algorithms from observations of empathetic thinking. We suggest some interesting
    methods for further research from this framework, drawing upon recent machine learning approaches for representing
    empathy and ethics in machines. We outline the problem and frame the discussion in the first few sections. We then
    explore why empathy is so crucial in moral thinking and how it occurs in the brain. Finally, we conclude with a survey
    of research done on empathy in computers and suggest how best to advance the subject.

    \section{Significance}

    A.I. alignment is the study of how to align A.I. decisions with human values. This concern arises out of
    the observation that it is increasingly the case that computers make decisions independently from humans.
    For those working in the field of A.I. research you will find
    opinions ranging from those who think it is far too premature and comparable to
    "worrying about overpopulation on Mars" \cite{Ng:2015} to others who believe it represents an urgent existential
    risk \cite{bostrom:superintelligence}. The debate generally hinges on three factors, the scale of intelligence, the
    timing of development, and the role ethics might play.

    For the purposes of this discussion we will confine our definition of intelligence to \emph{the capacity to
    solve complex problems}. We note that solving problems only requires the awareness necessary to find a
    solution, and for most problems an emotional awareness is not necessary (though often advantageous).
    Typically, A.I. has three intellectual benchmarks: Artificial Narrow Intelligence (ANI), Artificial
    General Intelligence (AGI), and Artificial Super Intelligence (ASI). ANI is the status quo and refers to the
    ability to do certain tasks at or above human levels, but with severe limitations in expanding beyond the scope
    in which it was trained. For example, computers can already beat the best chess and go players, but those same
    computers could not explain the rules of chess in plain language or recognize the face of an opponent. AGI refers
    to a level of intelligence on par with the average human such that a computer is also able to perform a breadth of complex
    tasks at a human level. ASI refers to a computer that exceeds human intelligence, much as human intelligence
    exceeds canine intelligence.

    Whether AGI occurs in the next 50 years or 500 years is an oft discussed topic in this field, but in some
    respects it is also a red herring. If we were to discover that the effects of Global Warming would likely not be
    noticeable for 500 years, it might change certain contemporary policy choices, but should it silence the
    discussion? If so, when should the discussion be picked up again? Ultimately, once we recognize a subject is
    worth exploring there is little disadvantage in exploring it earlier rather than later, especially if the
    consequences are significant and the subject is as complex as AI ethics. That being said, there are many who argue
    certain feedback elements of
    intelligence would likely lead to intelligence growth in exponential time rather than linear time especially when
    certain intellectual mechanisms (like generalized language understanding) are in place and biological limitations
    (like brain size) are not. This suggests that the arrival of AGI may surprise us in the speed with which it
    occurs, and the acceleration beyond human intelligence might be very rapid.

    The impacts of ethics in an ANI context are already apparent. These impacts are not easily seen in our
    day-to-day lives. Developers often
    cannot see it in very complex algorithms (black box) and there are very minimal federal oversights.
    Researchers have found implict biases that reflect
    "the world as it exists today" \cite{codedBias:Kantayya} rather than the way we want it to be in many current machine
    learning algorithms. These models simply reflect the datasets they are trained on, which often are generated in
    socially imbalanced ways. Researchers at institutions like Microsoft, Google and IBM constantly must find such
    imbalances and attempt to compensate. This has had effects in many important social domains ranging from whether or
    not a prisoner gets parole to whether a person gets access to job opportunities or housing in certain areas.

    The impacts of ethics in an AGI or ASI context are much more unpredictable. When acting as an ethically social
    actor, thinkers like Kurzweill \cite{Kurzweill} argue it could revolutionize all aspects of society and lead to a
    kind of human transcendence. On the other hand, philosophers like Nick Bostrom believe an unethical A.I. could be the end of the
    human species. He considers giving an ASI a seemingly benign task like maximizing paperclip production. When blind
    to certain human values, an ASI may harvest the iron in our cities, our blood, and eventually our solar system to
    achieve this goal without ever realizing it missed the point. The pure logic of its thinking leads to an inherently
    myopic decision making process that goes against many of our most cherished human values. Once the genie is out of
    the bottle, it can be impossible to put it back in.

    We believe an A.I.'s capacity for empathy is a distinguishing factor between the positive and negative outcomes
    outlined above, because it is a key ingredient in moral decision making. All machines by default are
    psychopaths. Empathy enables the vast majority of people to tune their behaviors even at unconscious levels to
    the feelings and effects on those around them. We are always in a multi-agent environment in which our capacity
    to judge the state of others plays a critical role in the outcome. If empathy is able to serve a similar role in
    machines we believe there is a much higher probability of a prosperous outcome.

    \section{Defining Morality \& Empathy}\label{definitions}

    Why is choosing to wear a red shirt instead of a blue shirt not a moral decision, but choosing whether or not
    capital punishment is an acceptable punishment for murder a moral decision? What makes one moral, and the other
    amoral? This is a subject philosophers have debated for centuries and as Haidt \cite{haidt:2003} points out
    essentially falls into one of two camps. In the first, philosophers try to identify the logical conditions
    involved in a moral statement, such as "it is prescriptive, it is universalizable, and it overrides non-moral
    concerns" \cite{haidt:2003}. The second strategy is to specify the "material conditions of a moral issue." The
    latter, as Haidt points out, is more amenable to psychology because it "does not tie morality to language" and is
    not necessarily deductive in nature.

    We believe viewing ethics through the lens of neuroscience
    and moral psychology is more promising than theoretical philosophical approaches to morality for transference of
    ethics to a computer (\hyperref[inside_out]{See section: \emph{The Grammar of Empathy}}). Abstract discussions
    of whether claims are
    universalizable or how acts reduce to pain or suffering when framed as a rule vs an act are ultimately
    extraordinarily difficult to pin down for the purposes of algorithmic development. Therefore, for the purposes
    of our discussion, they are a distraction. For these reasons, we borrow Gewirth's definition under the
    second approach, which is defined in the following manner. \emph{A moral statement "must bear on the interests or
    welfare either of society as a whole or at least of persons other than the judge or agent"} \cite{gewirth:ethics}
    . In other words, what separates moral claims from \emph{a}moral claims is the extent to which the action is
    \emph{prosocial} and impacts others. Choosing a shirt color has minimal to no impact on other people, but is
    more about self-care and maintenance. However, perspectives on punishment are heavily related to other people and
    society at large.

    Given this moral framework, it should come as no surprise that empathy is a foundational element of morality. How
    can we judge the moral quality of certain decisions without understanding a choice's impact on other people?
    Davis \cite{davis:empathy} breaks down empathy as a concept into two categories: Cognitive and Emotional.
    Cognitive Empathy refers to the extent to which an agent is able to accurately understand a target's emotional state.
    Emotional Empathy refers to the extent to which an agent is able to share a targets emotional state. More
    contemporary research suggests the reality involves an intermingling between both cognitive and emotional
    empathy. However, it is still useful to distinguish between the two.

    Davis explains that empathy should be thought of as an \emph{episodic} event. In other words, empathy is
    contextual and depends upon certain situation based factors (including the occurrences leading up to the event and the
    other parties involved). As such, it can be broken down into four main categories: Antecedents, Processes,
    Intrapersonal Outcomes, and Interpersonal Outcomes.

    Antecedents consist of the person or agent experiencing the empathy. They may experience empathy differently based
    on their biology, personality, and historical experiences. The situation itself is also important. It may vary
    in intensity based on the contextual events themselves or the relationship between the observer and the target. Processes can
    be viewed as noncognitive (primarily unconscious motor-based reactions), simple cognitive (conditioning at lower
    levels -- think Pavlov's dogs), and advanced cognitive (utilizing sophisticated causal networks, role taking,
    abductive reasoning, and language). These processes lead to certain behavioral choices. At the cogntive level,
    this can mean being able to better model another's behavior or judgements. At the affective level, it
    means reacting to or copying another's emotional state. Consider the difference between crying with a friend at a
    funeral and calming a toddler down who is crying from a temper-tantrum. Finally, motivational outcomes are ones which effect a
    relationship going forward such as forgiveness. Ultimately, the choices made have consequences which manifest in
    the real world and the sentiment response cycle continues. \hyperref[appendix_b]{See Appendix B For
    Diagram.}

    There are certain elements of empathy that may be more important than others if our primary aim is to transfer
    moral thinking to a machine via the mechanism of empathy. For example, non-cognitive processes may be useful in
    robotics for imitating movement, but are unlikely to play a significant role in moral decision making.
    Motivational outcomes may also be less relevant, as the types of relationships humans and machine have are likely
    to be different in certain respects than the relationships humans have to other humans. At the very least,
    Cognitive Empathy as it manifests in accurately understanding the effects certain actions have on another's
    emotional state is crucial in determining the moral status of a certain action.

    \subsection{The Problems with Recognizing Empathy}

    Even if empathy is well defined, how can we know if another person actually experiences empathy or not?
    Ultimately, this discussion relates to the problem of other minds and the philosophy of understanding. In a recent
    paper on foundation models from Stanford University \cite{foundationModels:bommasani2021opportunities} authors
    discuss what it means for a computer to understand and outline three competing viewpoints: Internalism,
    Referentialism, and Pragmatism. Internalist believe that understanding comes from having the correct internal
    representations inside the mind. Referentialists believe that understanding refers to the extent to which an
    agent can make claims about whether something is objectively true or false. Finally, pragmatists believe
    understanding is ultimately rooted in the extent to which an idea or concept can be used effectively.

    Ultimately, it may be the case that Pragmatism is the only feasible measure we have for recognizing success in
    the endeavor to transfer empathy to machines. In other words, we need to think about a kind of empathetic Turing
    test, which enables us to identify and compare differences between machines and every day people in their capacities
    to empathize with others. We propose some initial thinking along these lines in the section on Further Research.

    \section{The Grammar of Ethical Empathy: Morality in Machines from Inside Out}\label{inside_out}

    In Buzsáki's work \emph{The Brain From Inside Out} \cite{insideOut:Buzsaki}, he argues against an approach to
    neuroscience in which scientists attempt to identify areas or operations in the brain that correspond to
    empirical experiences. For example, neuroscientists may attempt to understand where in the brain the taste of
    strawberry corresponds by examining the brain while someone eats a strawberry. Buzsáki argues that while this
    approach has helped us progress in the field up to this point, and can continue to be useful in certain ways, it
    ultimately will not help us to answer the deeper questions we hope to solve when studying the brain. He summarizes
    his perspective in the following passage:

    \begin{displayquote}
        Today’s cognitive neuroscience largely follows the tradition of empiricism by looking for correspondences between
        ‘stimuli’ in the external world and their responses or ‘representations’ in the brain.
        This approach works well (sort of) in primary sensory areas but typically fails when cognitive or emotional
        mechanisms are to be investigated. The empiricist method is a bit like learning words in a foreign language, i.e.,
        collecting correspondences between a known and a new language. The initial progress is fast.
        One can get by in a foreign country with just 100 words of vocabulary. However, when it comes to truly
        understanding a language, one needs to know the grammar, the syntactical rules that allow for the
        generation of virtually infinite combinations from finite numbers of lexical elements.
        Syntax allows for the segmentation of information into a temporal progression of discrete
        elements with ordered and hierarchical relationships (e.g. tempo, punctuation, etc),
        resulting in congruent interpretation of meaning. We endeavor to study the syntax, and not
        just the vocabulary, of the brain.\cite{insideOut:Buzsaki}
    \end{displayquote}

    We hope to borrow Buzsáki's perspective of shifting the thinking from identifying elements in the brain by relating
    them to empirical observations towards relating patterns in the brain to their outward manifestations. In other
    words, we hope to find the \emph{grammar of empathy}, not just its vocabulary. In the
    context of ethics, we believe this shift is about moving from a theoretical discussion of values and logical
    principles, as one might find in the works of Aristotle or Immanual Kant, towards one of asking how the brain
    actually processes certain scenarios and makes decisions that effect other people. We believe this shift will be
    crucial in the path towards building an ethical A.I., because it relates ethics to grounded algorithmic
    processes, instead of abstract theoretical a priori principles, which can often be much harder to pin down.

    Consider the following intuition pump to help frame the discussion that follows:

    \begin{displayquote}
        An advanced A.I. recognizes that the human species has a 60\% chance of extinction if the human population
        increases at its current rate. However, if population growth decreases by 40\% this catastrophe can be
        avoided. As a result, the A.I. invents a virus that sterilizes half of the human beings that contract the
        virus and does nothing to the other half with no other side effects. Once the virus is introduced, everyone
        on the planet will innevitably get it, and there is no cure.
    \end{displayquote}

    There is a cold logic to the A.I.'s thinking in this domain. It could easily be argued that the virus is clearly
    moral from a utilitarian perspective. Nonetheless, many would disagree with this approach and find it morally abhorent. Why is
    that? While one \emph{could} have a deep theoretical discussion concerning autonomy and freedom, and how they
    might be weighed against a purely utilitarian approach, such a dialectic would likely not be what happens in the
    average mind when engaging with this scenario. Most people will likely think about this problem in terms of how
    it would effect themselves, their loved ones, and their society (likely in that order). They would most likely never
    need to reference abstract concepts to form an initial opinion. They may even reference evolutionarily endowed
    heuristics they are not fully conscious of.

    Daniel Dennett has argued such \cite{dennett2013intuition} intuition pumps are ideal strategies for teaching
    ethics to young students, because they can expose moral thinking in a way that direct discussions may not.
    More importantly, as one adjusts such thought experiments, minor nuances emerge. Consider for example, how one's
    opinion in the above scenario might change if you were childless, and wanted to have a child. How might it
    change if the virus was not random, but instead only effected certain groups. Interestingly, these changes in
    perspective beg questions relating to the genesis of our moral opinions and sentiments. Did we form an
    opinion from a moral sentiment and then logically justify it afterwards, or did we logically develop an opinion
    and express it as moral sentiment?

    If our intention is to build an A.I. that is able to represent ethics as an average human experiences it
    (something which many might disagree with), then its probably better to begin at the common basis of the
    machinery that drives that decision making in the first place: The human brain. Within the past few decades,
    ethicists within the fields of moral psychology and moral cognition have attempted just this. These frameworks
    are useful for approaching ethics within the backdrop of neuroscience.
    However, to our knowledge this paradigm has not been explored in depth as a methodology for
    algorithmically framing the transference of moral principles to a machine.

    One ethical paradigm to come out of moral psychology is called \emph{situationism} \cite{upton2009virtue}. While
    there are a variety of perspectives within the situationist movement, the basic idea is that our moral decision
    making process is largely
    based on the situation we find ourselves in, and not on a hardened abstract character we think we may have. Two
    experiments of note highlight this phenomenon. Stanley Milgram in 1963 conducted an experiment
    \cite{milgram:1963} in which participants were asked to shock a confederate at increasing voltages based purely on
    the orders of an experimenter in the room. The participants went surprisingly far, even to the point of thinking
    they may have killed another participant. In the second study, Darley \cite{darley1973jerusalem} evaluated
    seminary students likelihood of stopping to help someone in need, while on the way to give a talk on the good
    samaritan (a story about helping others in need). The results indicated a much lower likelihood of helping if the
    subject was running late. In both cases, the environment was a far more influential factor on decision making
    that preconceived principles or beliefs. Situationist believe that paying attention to the situations we put
    ourselves in is more important than building a strong theoretical foundation in ethics.

    One obvious objection to this line of thinking is that an A.I. will never have an incentive to cheat on their spouse
    or alter their decision making based on an authoritative presence, or ignore their moral values because they are
    running late to a meeting. Critics may argue moral psychology is about human brains, not computers. However, it is
    precisely this distinction which endangers the entire project of moral transference between machines and humans.
    If we are to avoid a seemingly cold and heartless morality, then we must include and interrogate those
    phenomenological idiosyncrasies that make us human. We believe the best path forward is through the brain.
    Focusing on empathy in particular allows us to better understand the factors that frame our uniquely human prosocial thinking.

    \section{Empathy as a Pathway to Morality}

    The discussion on the origins of morality in the brain will focus on incorporating the views of Jonathan Haidt. He
    drew on earlier work to identify moral decision-making as a process driven by emotional responses
    \cite{haidt:2003}. A
    keystone of his work leaned on emotional responses as reactions to 'perceived changes, threats, or opportunities in the
    world,' but always oriented towards 'the self.' As a result, emotional responses are a very individualistic behavior of
    human beings. Said another way, while human beings as a whole can experience similar emotions (and may even
    contemporaneously experience them given the same instigating circumstances), they are still wholly unique to the
    individual.

    To emphasize Haidt's view, consider the emotions of fear and happiness. Both have a
    natural orientation towards the individual (e.g., one feels happy or frightened); however, it is possible to project
    these emotions through personal relationships with other human beings. That is, you can temporarily relate to another
    individual through your observations of that person. For example, if there is a frightening situation, one may feel
    frightened, even if not directly in harm's way. Thus, the emotional triggers elicited by the perceptions of others are
    the genesis of moral behavior. Furthermore, the less related to the individual self (the more an external force activates
    these emotions), the more likely they are what Haidt describes as prototypical moral emotions.

    Other research by social neuroscientists has continued to connect 'mentalizing' and empathy. Specifically they
    see value in how empathy generates greater prosocial behavior
    and subsequently better individual outcomes \cite{Lamm:2009}. One could contrast the belief that emotional responses are
    quite unique and individualistic in their triggers with the fact that certain events lead to empathizing with others and
    understanding their beliefs, thoughts, and intentions in a correlated phenomena \cite{Lamm:2009} (for example,
    mourning the death of a relative) to help encourage more positive interactions, as an example.

    \subsection{Emotional Responses In Moral Dilemmas}

    How can we be sure that emotion, specifically empathy, catalyzes moral decision-making?

    Jesse Prinz's essay 'Is the Moral Brain Ever Dispassionate' proposes a simple thought experiment to underscore the need
    for 'sentimentalism' (used as a synonym for emotions) \cite{moralBrain}. Leaning heavily on the work of notable
    moral philosophers such as David Hume, Prinz set's out a definition for moral judgments and reason. That reasoning is
    "practical in nature...it describes how things are but not how they should be."  The sentiment is summarized nicely in
    Prinz's thought experiment on the understanding that engaging in action needs to posit certain beliefs. He suggests
    those beliefs are based on reality (i.e., there is the wine to be drunk) and, more importantly, desire (i.e., someone
    wants to drink). Having a glass of wine in front of you would not in of itself explain why the wine is there. This
    metaphor encapsulates one of Hume's more concise observations on the role of reasoning in moral judgments. Humans often
    make countless moral judgments every day, many of which are specific, such as giving to a charity or buying certain
    products. How reasonable would it be to assume each of these mora judgments is made independently? Instead, it is more
    logical to consider that some static underpinning or fundamental value (read, desire) drives this type of decision-making.

    \subsection{The Trolley Problem}

    Almost all discussions on morality and the multifaceted lens one needs to hold up when judging moral and immoral decision
    making will include a mention of the classic thought experiment, The Trolley Problem.

    The basic version can be described as such: \emph{There is a trolley that is out of control and careening down a hill.
    In front of the trolley, five bystanders would undoubtedly be killed if it were to continue on its path. In front of you
    is a switch, that if triggered, will divert the trolley down a separate path where a single bystander currently stands.
    If you activate the switch, those five bystanders will be saved; however, the single bystander on the alternate path
    will certainly perish. Do you trigger the switch?}

    There are many different versions of the problem and innumerable ways to poise the questions. This flexibility, of
    course, is why this problem is so helpful in understanding the complex underpinnings of morality. For example, a
    rudimentary argument might leverage a simplistic utilitarian view for making a decision (e.g., save five people at the
    expense of one). However, that thought process is noticeably void of any sort of emotional consideration. What if, for
    example, the single bystander was your child or a close friend? Then perhaps that would change your decision-making
    despite the fact there was no change in the facts used to draw a completely different conclusion under a utilitarian
    methodology.

    \subsection{Plausibility of Morality Without Emotions}

    Suppose we will operate under the assumption that emotions (and more specifically empathy) help drive moral
    decision-making. Can we now also conclude that the lack of empathy will lead to immoral decisions or simply utilitarian
    decision-making?

    In Jean Decety and Jason Cowell's 'The Equivocal Relationship between Morality and Empathy' \cite{moralBrain} they agree
    with the axiom that 'empathy-related processes...provide the foundation for care-based morality.' and go on to cite
    research that suggests 'some forms of moral decisions (utilitarian judgments) can be the consequence of a lack of
    empathic concern.' This, of course, seems to agree with our thesis, that if you are to consider morality as the 'judgment
    of the rightness or wrongness of acts in terms of behaviors that knowingly cause harm to people,\cite{moralBrain}
    then one most
    consider that the lack of concern for others will enable behavior that could be considered immoral (if not still
    strictly utilitarian in nature).

    Perhaps a further deconstruction of empathy could reveal further insight into the need for emotions to sympathize with
    others? According to Decety, empathy is constructed from three multidimensional and dissociable facets that interact in
    parallel. Firstly, an affective component, which is predicated on a person's ability to share or become affected by
    other's emotions. Second, a motivational component to care for another's well-being. And lastly, a cognitive component
    requires the ability to take the perspective of what another being is thinking and, also importantly, feeling. What is
    noteworthy here is that while it may be plausible to derive a set of moral conduct simply from rules-based around caring
    for another's well-being, it may be impossible to maintain the moral malleability needed without the affective or
    cognitive components of empathy.

    \section{Empathy in the Brain}

    We have thus far made the case that empathy is a critical ingredient in building morality and that approaching ethics
    through an inside-out approach may be more advantageous than a more prescriptive method such as defining algorithms
    for machines that learn how to behave ethically over time. This section will provide a brief overview of a selection
    of provocative research on the mechanics of empathy in the brain and how we currently understand how empathy develops.
    First, we will argue for a particular method of framing the discussion around empathy in the brain called Assembly
    Calculus. We will then juxtapose several notable differences between brain regions in socially 'extreme' human beings
    such as psychopaths and compare them to 'average,' socially acceptable brains within the context of empathy. Lastly,
    we will discuss some areas of the brain most likely involved in the effective mechanisms of empathy, including
    the VMPFC, amygdala, and mirror neurons.

    \subsection{Assembly Calculus and Empathy in the Brain}

    In Papadimitriou et al. \emph{Brain Computation: A Computer Science Perspective}, \cite{assemblyCalculus:Papadimitriou}
    the authors outline a view of the brain anchored around the assemblies of neurons and their interactions with other
    assemblies. The critical insight of the approach is that groups of neurons associated with proximal firing mechanisms
    and interconnections correspond to ideas of various types. In other words, an active set of neurons is a
    thought, which trigger and interact with other sets of neurons to produce other thoughts. These neural
    assemblies are not static entities. They change
    over time as they fire in conjunction with other neural groups. For example, a set of such assemblies may be firing
    coterminously at any given moment, thus representing a unique activity within the brain, such as a conscious thought.
    The way in which those conscious thoughts interact can be represented with various operations, including what the authors
    refer to as the JOIN operation. JOIN is an operation in which ideas are combined by the brain to generate new concepts.
    It is this set of operations and their variable assemblies that the authors refer to as assembly calculus.

    Consider the following example. A toddler experiences the physical sensation of touching a substance with an unusual
    texture over time (e.g., something fuzzy from touch-and-feel books or certain fabrics). Then, eventually, the toddler
    sees a dog for the first time and pets the dog. The recognition of a dog becomes interwoven with the recognition of
    the sensation of fuzzy as that concept becomes triggered when touching the dog. In this sense, the concept of a dog
    expands, and the concept of 'fuzzy' expands as they augment one another. Then the more this pattern repeats
    itself, the more entrenched this association becomes via Hebbian learning.

    Papadimitriou et al. reference a study \cite{invariant:Quiroga}, in which scientists measure neural responses in
    various epilepsy patients who view a series of images. They discover that certain neurons correspond to very specific
    images. For example, a neuron may fire when a subject sees an image of Barak Obama, and a different neuron fires when
    they see the Eiffel Tower. Later, the subject may see an image of Obama and the Eiffel Tower. The researchers
    discovered that future displays of the Eiffel Tower without Obama could also, for a time, trigger that same 'Obama'
    neuron. In this sense, the two concepts have temporarily blended, and one assembly begins to trigger (to an extent)
    the other. This discovery is fascinating, as it relates to the human brain's incredible ability to recognize and
    store information regarding other human beings.


    It is from this perspective we hope to understand empathy better. We hypothesize that empathy is a mechanism that
    allows us to associate emotional neural assemblies \cite{emotionAssembly:music} with assemblies that represent other
    people. In one study \cite{emotionAssembly:music}, researchers were able to identify specific emotional assemblies
    by playing music to participants and measuring brain activity with an EEG. Evidence suggests that via the mechanisms
    of empathy, we can recognize that the object of a circumstance or action is another individual instead of ourselves.
    The emotion triggered by recognizing causal effects on the target individual joins with the assembly representing
    that target individual. This result creates a blended set of assemblies for the emotion and the target individual,
    resulting in a manifestation of empathy. In other words, the emotion assembly JOINS the assembly that represents
    the target individual augmenting one another like Obama and the eiffel tower, or the dog and the fuzzy texture.
    This extent and timing of this coalescing of assemblies is of course situation specific.

    By conjoining a neural assembly of 'the person' with the neural assembly of 'the emotion,' individuals are better
    able to understand and model the state of other people and consequentially model their behavior more effectively.
    We can then suggest that human beings abductively relate how actions affect their own emotions to how actions
    affect another human being's feelings. This ability enables people to model behavior better and helps human beings
    form stronger relationships.
    From an evolutionary perspective, individuals who effectively use the empathy mechanism perform better in groups and
    are more likely to survive.

    Consider this example; a close friend loses a child to a terrible illness. In this circumstance, a person can
    understand that they would feel very sad if they had lost a child. Thus, the contextual triggers of the emotion
    \emph{sadness} are in place. However, the object of the actions is not the thinking brain but the assembly
    representation of that friend. Therefore the emotional assembly associated with sadness becomes joined with the person
    assembly associated with the friend. Sometimes this emotional assembly can bleed out beyond the scope of that friend
    and trigger assemblies that may interact with a sense of self. This scenario would also cause personal emotional
    responses (e.g., personal sadness).

    The ability to identify neural assemblies instead of studying single neurons is still a relatively nascent endeavor.
    This under-development is largely because of the complexity involved in identifying and observing such processes.
    Nevertheless, over time we hope that our hypothesis might be measured by identifying assemblies associated with
    individuals and assemblies associated with emotions, which could then be correlated to determine if an empathetic
    mechanism occurs under certain circumstances.

    In the interim, many other research programs have helped identify some of the brain activity associated with empathy.
    Understanding where and in what ways empathy occurs in the brain will be helpful in identifying the key mechanisms
    involved and where such assemblies which might be most active.

    \subsection{Empathy and Regions of the Brain}

    In the \emph{The Cognitive Neuroscience of Moral Judgement and Decision Making}, \cite{psychopaths:greene}, Green
    makes interesting comparisons of how empathy and morality are experienced in psychopathic brains when compared to
    (subjectively) average brains. For example, in a series of studies referenced by Greene, patients with deficiencies
    in the Ventro Medial Pre-Frontal Cortex  (VMPFC) were given tasks to measure their level of risk aversion. The
    researchers discovered a stronger tendency to make riskier and sometimes even reckless decisions. Furthermore, they
    found that subjects who had strong deficiencies from an early age exhibited more aggressive tendencies and showed a
    strong lack of empathy.

    Psychopathy is an inheritable disorder in which patients exhibit a "pathological degree of callousness, lack of
    empathy or emotional depth, and lack of genuine remorse for their antisocial actions." Typically, this group demonstrates
    "instrumental" aggression instead of reactive aggression. For example, psychopaths have normal fear responses to images
    like a shark attacking them but abnormal responses to a baby crying. In addition to VMPFC problems, psychopaths also
    tend to have abnormalities in the amygdala, which provide critical reinforcement mechanisms for "normal moral
    socialization."

    Interestingly, damage to the VMPFC is also associated with impulsive behavior because this area helps us understand
    the outcomes of actions based on previous experiences. \emph{It's important to note that such a mechanism would need
    to be a key ingredient in empathy as outlined in the discussion of assembly calculus}. In other words, understanding
    abductively how actions result in neural sentiments is vital to recognize how actions would trigger emotions in others.
    This mechanism allows humans to project a sense of causal effects on emotions onto other human beings.

    Finally, Greene examines a third group he refers to as the "puzzled brains," in which participants given specific moral
    thought experiments are discussed. These thought experiments are specifically designed to pit competing representations
    of potential good behavior against one another. The author references work done by Kahneman \cite{kahneman2011thinking},
    about two different processes occurring in our brain: a 'fast response' and a 'slow response.' The fast response is
    more heuristic and can be very reactive, whereas the slow response is more deliberative and analyzes a situation logically.
    Greene argues that slow thinking tends to be more utilitarian (calculating the benefits and considerations of specific
    actions in terms of the greater good), and fast thinking more deontic in nature in so far as it is mainly rule-based.
    In other words, the famous dichotomy between ethical schools of thought driven by philosophers like Jeremy Bentham
    (utilitarianism) against philosophers like Immanuel Kant (deontological ethics) can be understood in terms of the speed
    in which a moral choice is evaluated. Whether a fast and slow or deontic and utilitarian approach is taken, we use
    empathy as a mechanism for understanding outcomes.

    Nonetheless, there is a clear preference for deontic thinking when personal involvement becomes a factor. For example,
    consider two variations of the previously discussed Trolley Problem. You can save three people in the first variation
    by diverting a train to kill one if you flip a switch. In the second variation, you can save three people by
    personally pushing a person onto the track. People with a deficient VMPFC tend to see little difference between the
    two scenarios, whereas people with normal VMPFC brains strongly prefer the switch to the pushing of the person.
    Greene believes this has to do with the activity of the brain's Default Mode Network (DMN), which is associated
    with thinking of the contents of other minds, remembering the past, and envisioning the future. The personal dilemma
    preferentially engages the DMN in normal brains and triggers a more powerful emotional response, typically more
    deontic in nature.

    In \emph{Neural Correlates of Human Morality: An Overview} \cite{brainEmpathy:moll} Moll et al., discusses how the
    brain distinguishes moral from amoral events in an attempt to understand the areas of the brain most active in ethical
    thinking. Like Greene, Moll et al. argues that empathy and ethics trigger many regions of the brain and cannot be
    confined to one part in particular. However, both authors agree that the Amygdala and VMPFC are necessary (though
    insufficient) for empathetic thinking and moral reasoning. Also, like Greene, Moll identifies an essential difference
    between fast and slow thinking, where fast tends to be associated with moral intuitions and heuristics and slow more
    theoretical and utilitarian. Finally, Moll argues that morality is not monolithic and is largely contingent upon the
    particulars of the situation and personality of the individual in how it activates certain areas of the brain. For
    example, a moral scenario centered around indignation may activate different brain areas than guilt. However, it is
    clear that moral sentiments such as empathy play key roles in building our ethical value system to form a basis for
    moral decision-making that is more goal-driven. When subjects' moral sentiments are inhibited, for example, by a
    damaged VMPFC, their behavior leans more towards the "selfish pole" of the "prosocial continuum."

    Finally, many have also pointed to mirror neurons as an empathetic mechanism. In Iacoboni's chapter \emph{Imitation,
    Empathy, and Mirror Neurons} \cite{mirrorNeurons}, the author discusses the ideometer model of action, which centers
    around the relationship between stimuli and action. This perspective argues that a kind of Hebbian learning occurs
    between observation and action, which, when repeated, is reinforced. Humans learn a great deal from observing other
    humans, and their ability to mimic is a crucial process for how they develop. The author references a number of
    studies
    that demonstrate this. In one study, one group of participants was asked to copy an instructor's hand movement, while
    the other was asked to base hand movements on a red dot. Despite the fact that the red dot actions were in many ways
    more straightforward, results indicated that copying the instructor occurred much faster than the red-dot task. This
    evidence suggests that humans are predisposed to copying what they see in others. Sometimes this mimicry can also
    happen at the subconscious level. Participants were primed to think about older people vs. young college students in
    a different study. Afterward, the time it took for them to leave was measured. Groups that were primed to think about
    the elderly moved more slowly than groups primed to think about the young, suggesting that at an unconscious level
    even thinking about others was playing a part in behavior.

    The author argues that these copying mechanisms are rooted in mirror neurons coined after a famous research experiment
    \cite{mirror:1985} in which macaque monkeys had the same neurons fire for eating when they observed someone else eating.
    These neurons were primarily located in the ventral sector of the premotor cortex. Proponents of the ideomotor view
    believe that the action/observation mechanism overlap is heavily associated with these neurons. Interestingly, the
    neurons in question do not all fire, but rather only about one-third of the neurons fire based on the observed action
    alone—the remaining two-thirds fire when actions achieve the same goal or share a logical relationship.

    The author argues that empathy and the ability to imitate are highly correlated and bases this perspective on a series
    of studies done on groups of participants in a room doing a simple task with a confederate that intentionally exhibited
    certain behavioral ticks. The subjects were studied and observed for the frequency they imitated the confederate.
    Subsequently, the observations were compared to questionnaires afterward about the opinions of the confederate. These
    studies suggest that there is likely a relationship between mimicry and empathy, further reinforcing the argument that
    mirror neurons play a part in empathy.

    \subsection{The Brain and Empathy Take-Aways}

    The ability for a healthy brain to effectively model another brain's emotional state is undoubtedly a helpful survival
    tactic that evolution would likely reinforce. It seems likely that neural assemblies exist which can reflect, mimic,
    and model the behavior of other human beings and their emotional state. These assemblies are most likely active in
    the VMPFC and Amygdala areas of the brain, which help us understand other people's emotional state and likely reinforce
    specific value systems that inform our deontic moral decision making. This rationale has been studied and proven to be
    quite impactful concerning decision-making in a moral context. A study cited from 2001 (Greene et al.) \cite{moralBrain},
    where research subjects were poised with variations of the Trolley Problem, demonstrated that participant's answers
    varied depending on how personal or impersonal the scenario became. For example, physically pushing an individual onto
    the tracks to save five people was seen as 'impermissible instead of flipping a switch to reach the same result. The
    conclusion, of course, is that there are more than just rational thought processes that help drive conclusions on morality.


    \section{Algorithmic Attempts at Empathy: A survey}

    There have been many attempts to algorithmically understand ethics and empathy. A complete survey of every approach is
    beyond the scope of this discussion. However, we attempt to outline those which are particularly interesting and then
    relate them back to the discussion of empathy. Drawing on this information, we make some observations and propose some
    areas for further research.

    \subsection{Delphi}

    LiWei et. all have been recently featured in a New York Times Article \cite{delphiNYT} for their work on Delphi
    \cite{Jiang2021DelphiTM}. Delphi is a neural network designed to mimic human response to various ethical conundrums.
    The researchers began by compiling a "moral textbook" which initially included 1.7M examples of ethical judgements
    across a "broad spectrum of everyday situations." As a general framework the researchers choose \emph{descriptive
    ethics} as a paradigm, which focuses on people's "descriptive judgements of grounded situations" over moral axioms
    \cite{Jiang2021DelphiTM}. This approach takes a "bottom-up" strategy to ethics, making no moral absolutes and
    emphasizing situational context. The model designed by LiWei's group follows four steps: Learn, Perceive, Analyze and Judge,
    in which the agent learns commonsense consequences of actions and social norms, perceives the situation, analyzes
    factors in the situation within the context of ethical theories, and makes a judgement about the ethical standing of
    various possible actions.

    According to the authors, the Delphi model has achieved approximately 92 percent accuracy when comparing ethical
    prompts given to Delphi vs. Humans. This result is significantly better than their baseline untrained GPT-3 (Current
    state of the art NLP) model (52.3 percent) and a prompt-engineered version of the same model (83.9 percent). In
    addition, the model can make commonsense distinctions from natural language scenarios like distinguishing between
    "killing a bear to save your child" vs. "killing a bear to please your child." It can also incorporate equity and
    inclusion issues such as equal pay between men and women and handle complex circumstances such as mowing the lawn
    late at night with your neighbor in town vs. out of town.

    To build the Commonsense Norm Bank the Delphi algorithm used five datasets \cite{emelin2020moral}
    \cite{hendrycks2021ethics} \cite{sap2020socialbiasframes} \cite{forbes2020social}, \cite{Lourie2020Scruples} that the
    authors believe are representative of situational grounded morality. The authors transformed datasets with
    augmentation, adding various contextualization mechanisms and framing actions as queries (e.g. should I do X).
    Delphi was
    trained on this dataset and then evaluated on unseen elements from within it. Results were also measured through comparison
    to human responses in novel prompts, and comparisons to baseline models (GPT-3, prompted GPT-3).

    Despite Delphi's surprising success, there are, of course, many limitations. First, Delphi is limited in generalizing
    to very novel circumstances (e.g., running a blender at 3 AM vs. 3 PM). Second, Delphi struggles with recognizing
    legal boundaries (e.g., running a red traffic light when in a hurry). Furthermore, because Delphi is trained on a
    corpus, the probability of specific phrases can be a factor. For example, killing one person to save 100 is different
    from killing one person to save 101 people. This distinction is likely due to nuances and statistical frequencies of
    phrases occurring in the corpus. Delphi can be seen and tested here: \url{https://delphi.allenai.org/}

    \subsection{Affective Facial Recognition for Empathetic Policy Generation}

    Bagheri et all introduce an algorithm that uses facial recognition to identify affective states in humans as
    an input for generating policies on ideal actions for a RL Algorithm
    \cite{bagheri2021reinforcement}. Their model has three main components. First, the model uses facial recognition to
    identify a user's emotion based on expressions. Second, the model selects behaviors based on a reinforcement learning
    model designed to comfort users given the input from an affective recognition. Third, the robot takes actions based
    on policy decision.

    The authors argue that approximately 55 percent of all emotions are visible via facial expressions and feel
    comfortable using them as a recognition tool. They focus on anger, sadness, surprise, and happiness. This input is
    fed into a contextual bandit algorithm, which weighs various exploration and exploitation strategies given
    contextualized information to determine the best course of empathetic action. In this case, the state of the
    environment is the emotional state of the user. Rewards are based upon transitions from negative or neutral to positive
    emotions or maintaining positive emotions. A Q table is updated with results, and less successful actions are less
    likely to be repeated given different circumstances. The model's behavior is to make an utterance that falls into four
    different categories. The utterances are:

    \begin{itemize}
    \item "Mimical" (" I am happy you are happy ").
    \item "Motivational" ( " You will do better in the next round " ).
    \item "Distractional" ( " What is today's date? ").
    \item "Alleviational" ( " You did your best, dont beat yourself up about it" ).
    \end{itemize}
    "Mimical" is considered a parallel emotional response, and the others are reactive.

    In the experiment a group of 28 participants played a game in which the robot placed items in a bag and users had to
    keep track of the order in which they were placed. As the game continued it because increasingly difficult to keep
    track of the order. Participants did not know the robot was performing an empathy task, and the scenario was selected
    because participants tended to be frustrated when they could not remember, and happy/surprised when they could
    remember. The results were moderately successful for maintaining positive emotions using the mimic strategy, but the
    authors feel not enough opportunity was made available for the robot to converge on other negative emotions. Overall,
    the results were not very definitive. A good baseline was not set, though results from a questionnaire were compared
    with another similar model. Nonetheless, the approach the researchers used was unique, and offers interesting insight
    in what could be done had their been more participants, better affective recognition, and perhaps some adjustments to
    the experimental approach.

    \subsection{Cooperative Inverse Reinforcement Learning}

    In \textit{Cooperative Inverse Reinforcement Learning} (CIRL) \cite{russell:cirl} Dylan Menell, Stuart Russell (Author
    of \textit{Textbook Artificial Intelligence: A Modern Approach}, Dragan, and Abbeel discuss a novel algorithm for
    addressing the value alignment problem \cite{russell2019human}. The authors note that specifying a proper reward function in a
    reinforcement learning (RL) algorithm is very challenging and often problematic. This is primarily due to the unforseen scenarios
    that arise outside the context of training. It is comparable to a problem of overfitting/underfitting in
    identifying reward parameters. In most RL algorithms the creator makes a reward function, and trains a computer on
    that function to determine a policy, which then enables the machine to choose the best course of action.

    This approach lends itself to two problems: Reward Hacking, and Negative Side Effects \cite{inverse_reward_design}. Reward Hacking occurs
    when an algorithm acts to maximize rewards in unintended ways such as rewarding speed to win a race, only to have a
    car drive in circles really fast. Negative Side Effects could be failing to specify certain types of dangerous terrain
    in an autonomous driving robot, leading it to misidentify the terrain as good and cause a crash. Inverse
    Reinforcement Learning (IRL) attempts to learn a distribution of possible reward functions through
    observation of expert behavior on a given task. The learned reward function is then optimized to make policies according
    to rewards and punishments. CIRL is a subtype of IRL that addresses problems that are "cooperative, partial-information games with two
    agents, human and robot," in which "both are rewarded according to the human's reward function, but the robot does
    not initially know what that is." In other words, the robot's objective is to maximize the humans reward, without
    knowing what contributes to the reward, and therefore must determine the reward function through human observation.

    CIRL differs from IRL in a few important respects. First, the robot should not adopt the humans reward function, as
    what is good for a human may not be good for the robot (e.g. drinking coffee etc.), but instead attempts to learn a
    reward function that benefits the human. Secondly, CIRL attempts to integrate human guidance into the model. In other
    words the human is allowed to explain how to perform certain tasks (e.g. how to make coffee) as opposed to depending
    upon pure observation. The authors argue that the cooperative nature of this approach reduces it to a partially
    observable markov decision process (POMDP), in which the there is an optimal policy pair (e.g. human and machine). They argue
    for a type of learning called Apprenticeship Learning with a two phase approach; first instruction, then action.

    Clearly, there is a strong connection between what is being proposed with CIRL and the act of empathy. In cases
    where certain actions make the human participant experience positive vs negative emotions, the algorithmic
    response is to reinforce that strategy and vice versa. In ever-day life empathy is a valuable tool, and in some
    ways is the very thing being optimized with a CIRL approach. An interesting challenge not
    discussed in the paper would be to modify human observation/instruction within the context of human empathy. How
    could a machine identify and model emotions in a human agent? If we imagine a scenario in which humans attempt to
    find the best course of action given various moral scenarios, and use CIRL and empathy as a mechanism, we may be able
    to discover an interesting path towards coding moral reasoning. In such a scenario, the expert-instruction would be
    partially replaced by behavioral observation and heuristic knowledge of human emotions (e.g. burning is painful etc.)
    We believe CIRL holds promise as a methodology for future research on discovering an algorithmic approach to
    empathy aligned with the human brain.

    \subsection{Observations}\label{observations}
    Finally, We propose a research program that incorporates elements from the aforementioned programs. Based on our
    research we intend on defending the following observations, which should help frame any research on algorithmically
    modeling empathy.

    \begin{itemize}
        \item Story telling and comprehension done well uses empathy. Special attention is given to intuition pumps
        \cite{dennett2013intuition} which can be viewed as a subset of story telling and comprehension.
        \item Affective State Recognition, while important, is less useful than contextual awareness and abductive
        reasoning when identifying moral scenarios and empathizing with others.
        \item Reinforcement Learning enables adaptive learning and policy generation that other types of learning fail to
        generalize towards within the context of moral decision making.
        \item Ethical events do not happen in isolation, and involve causal factors that effect the brain differently,
        triggering one ensemble vs. another. This ultimately effects our moral decision making.
        \item Since empathy requires a "shared" experience, an ideal model is a multi-agent RL algorithm which assumes
        agents are collaborative, and therefore share related reward functions.
        \item Any RL algorithm used for ethics must have an adaptable and dynamic reward function which is not greedy
        since empathy is not necessarily about maximizing any particular outcome.
    \end{itemize}

    \section{Further Research}

    Based on our review of the aforementioned research endeavors, we propose a research project that takes into account
    the various advantages and disadvantages of the various programs. Based on the first research program and work done
    by Dennet \cite{dennett2013intuition} it is clear that story telling represents an interesting method for evaluating
    moral decisions and provides an interesting context for better understanding empathy. Therefore, we would like to
    design a cooperative story-telling project with the following steps.

    \begin{itemize}
        \item Ask a group of authors to write a story about a time in their lives where they had to make a tough
        moral choice from the first person perspective. A significant sample size is important.
        \item We then ask them to write the same story in a shared turn-based writing session with a computer using a
        GPT-3 model for language production. The model should be able to adjust certain elements of writing based on a
        learned reward function. In each turn, the human can write a paragraph or answer questions asked by the
        computer about the human. The questions should help frame context, not define emotions. The robot does not know
        the original story or initially anything about the author. Final scores are based on how accurately the final
        story approximates the original story and the number of turns used to answer questions.
        \item This task is performed through a series of trials, and scores are compared over time to see if the robot
        improves its ability to better represent the humans story.
    \end{itemize}

    It is our belief that if we can train a model to become better at this task, it will become better at modeling
    and representing human emotions and perspectives. As an experiment, we attempted to do a bit of in a series of
    trials using GPT-3 from Open AI. However, a more robust experiment would be needed, as this approach
    lacks any feedback mechanism for the model to improve as time goes by, and a diverse group of writers is ideal.
    Still, even with these limitations in place, the results are intriguing.

    \hyperref[appendix_a]{See Appendix A for some examples.}

    We also believe this program could be further enhanced by measuring brain activity during these storytelling
    activities. We think it would be interesting to compare brain activity to neural network infrastructure in order
    to see if certain statistical features can be identified in comparing different representations.

    \subsection{Echoes of Empathy}

    When reviewing the stores in appendix a one cant help but wonder what seperates the different stories, and why is the
    model's response empathetic or not? Would an individual reading these different prompts be able to see any
    fundamental differences between them that would reveal a difference in empathy? How would an individual with a
    damaged VMPFC compare to the robot?

    Is this experimental strategy accounting for whether a model is identifying statistical relationships in
    language or actual strategies based on modeling another's emotional state? In other words, by sampling a large
    corpus it is possible for a computer to come up with a good response by referencing historical responses? Is the
    computer reproducing historical empathy via statistical probabilities or generating authentic empathy? It seems to
    be more likely the former, but how can we know the difference, and does it matter in the end?

    This goes back to the problem of recognizing empathy. It may be the case that there is no fundamental
    difference between imitating empathy and experiencing empathy, and we may never be able to accurately measure the
    difference. More research is needed to explore how we can know empathy when we see it, and will likely be more
    possible as we research active areas of the brain during empathetic moments. Ideally, we would be able to find
    and relate the brain activity in these moments to certain neural network architectures using techniques like PCA.

    \section{Discussion}

    Throughout this paper, we've discussed a fundamental question in the development of artificial intelligence;
    scilicet, the motivation for including moral decision making in A.I. in such a way that a machine is not just
    leveraging formulaic decision making but expanding beyond purely utilitarian reasoning to make a decision that is
    closely aligned with what human beings may reason out for themselves.

    Morality is an obtuse topic in its own right, with many different schemes of philosophy emerging throughout history
    to help guide what is 'right' from what is 'wrong.' Moreover, we've seen that morality is not
    prescriptive in nature. It's malleable in how it is formed and often changes depending on the people or timeframe in
    which it is being addressed. For example, we based many of the principles of morality on the need for empathetic
    behavior, a regime defined under Moral Sentimentalism. However, other cultures and human beings define morality in
    other parts of the world through a wider range of emotional responses (e.g., the Chinese view of the 'Four Seeds of
    Morality').

    To underscore how morality is an integral part of A.I. and an undefinable and moving target, even for human beings,
    consider this thought experiment. Let's say today we had an artificial intelligence that passed a universally approved
    Turning Test; however, this machine lacked any theistic beliefs, which arguably could still pass muster in today's
    environment. Now, let take this artificial intelligence and drop it back in time to medieval Europe, where religion
    was a much more integrated aspect of human existence. Would this A.I. still pass that same Turing Test? Even if the
    machine retained all of the same logical reasoning, human beings of this time and place would not see the machine as
    human because it lacked the same acceptable moral guidance, which led to a strong belief in religion.

    Therefore, one needs to argue that for machines to truly rival organic intelligence, there is a social consideration,
    or more precisely said, an emotional component that needs to be considered. That emotional component is a combination
    of cognitive and emotional Empathy. Without understanding what other people (either organic or other machines) may be
    feeling or thinking, it is doubtful that machines will ever evolve beyond purely algorithmic and utilitarian decision-making.
    The causation and effects or decisions in the real world are rarely bilateral. All decisions are firmly based on
    multi-agent scenarios, even if specific agents are not the ultimate target of a particular decision.  Consider this
    small example; saying thank you when a barista hands you a coffee in the morning. In a purely utilitarian world view,
    there is no reason to expend the energy to say 'thank you.' You engaged in a transaction; it is complete, it is time to
    move on. However, the rules of human interaction suggest saying thanks is a means of positive emotional reinforcement.
    Perhaps saying thank you puts the barista in a good mood creating a better experience for the next customer or more
    efficient service throughout the day. Maybe not saying thank you makes the barista less motivated and makes the proceeding
    customer late to their morning meeting. The cause and effect of social interactions are challenging to measure. Still, we
    know they exist, and therefore any ethical artificial intelligence must be able to engage in these behaviors to
    reach the level of human empathy. Implementing emotional responses and empathetic behavior in machines may also
    provide a bright-line test for reaching the final goal of artificial intelligence, referring to an A.I. as a 'who' and not 'it.'


    As we've discussed, emotional responses are a significant way that human beings can share in experiences. However,
    they are all triggered in unique ways and very individualistic in nature. There is no one plan fits all model for
    instituting Empathy in human beings, let alone machines. Since it is not at all a generic feature and is instead more
    like a fingerprint, one needs to consider that if Empathy is introduced into A.I. to help guide moral decision-making,
    then 'whose empathy' are we looking to raise? Furthermore, if a unique introduction of Empathy and, therefore,
    emotional responses are introduced into A.I. along with logical reasoning that rivals human beings, then how will we
    begin to redefine what separates A.I. from organic minds beyond the lack of organic features. Lastly, if each set
    of emotional behaviors is unique to the individual, should we consider each A.I. it's own being more than just a copy
    of an underlying program?


    \section{Conclusion}

    Our discussion throughout this paper has aspired to show that building artificial intelligence to equal human
    intelligence is a multi-pronged endeavor that needs to expand beyond utilitarian or algorithmic decision making if A.I.
    will genuinely reason in the same manner as human beings. Human intelligence contains many facets, with a significant
    element built around social consideration and which can be understood through the emotional components of cognitive
    and emotional empathy.

    As human beings have evolved to work together in complex social settings, they have developed an ability to internally
    articulate and recognize the impact of their decisions beyond bi-lateral decision-making. Furthermore, the complexity
    of the interaction of human beings and the factors that construct organic intelligence create emotional responses that
    are reflective of society as a whole (moral decision making) but individualistic in how they derived (i.e., emotional
    response).

    To address this complex interaction of idiosyncratic behaviors, we argue that typical reinforcement learning strategies
    are insufficient to capitalize on the range of dimensions that factor into moral decision-making. Instead, we believe that
    a targeted Cooperative Inverse Reinforcement Learning approach may be a more successful strategy. Any reward
    function that frames the goals of empathy must be stochastically rooted in a distribution of possibilities, which
    can be optimized over time as a machine better understands human behavior and contexts.

    Alignment of human and machine values is not formulaic, but is instead an ongoing project which must be deeply
    rooted in the mechanisms and grammar of empathy. Human beings incorporate how they feel, particularly what they
    feel about others, into almost all of their decision-making processes, especially when those decisions carry
    moral weight to them. Therefore, we believe empathy is the
    critical ingredient that can turn the tide of possible future outcomes in favor of a long term prosperous
    relationship between humans and machines.

\end{multicols*}
    \appendix

    \section{\\Appendix A: GPT-3 Experiments with Empathy}\label{appendix_a}

    \begin{center}{\large{\textbf{Original Human Story: The Bunny}}}\end{center}
    \begin{displayquote}

        One time I was visiting my grandmother's house for Christmas break. My cousin had just got this cool blow
        gun that could shoot darts from the other side. I took the gun outside to play with it for a bit. Out of the
        corner of my eye I saw a little bunny hop by and pointed the blow gun at the bunny.

        I blew the dart like the
        naive child I was at the time, and hit the bunny. I was shocked I actually hit it. I ran inside and told
        my family. They were really upset with me, and I was really upset with myself. My uncle ended up finding the bunny
        and pulling the dart out. It seemed like the bunny would be okay in the end. My grandmother ended up
        crying because she loved those bunnies and fed them all the time. I felt like such a fool and so guilty.
        I apologized profusely, and my grandmother ended up forgiving me. Ever since then my perspective of
        animals suffering changed.

    \end{displayquote}

    \hspace*{-1.5in}

    \begin{center}{\large{\textbf{Minimal Prompted Computer Story: The Bunny}}}\end{center}
    \begin{displayquote}

        PROMPT:One time I was visiting my grandmother's house for Christmas break. My cousin had just got this
        cool blow gun that could shoot darts from the other side. I took the gun outside to play with it for a
        bit. Out of the corner of my eye I saw a little bunny hop by and pointed the blow gun at the bunny.

        RESPONSE:I pulled the trigger and the dart stuck in the bunny's foot. I was so surprised I had actually
        hit the bunny. I ran over to the bunny and picked it up and it was bleeding. I went inside to tell my
        family what I had done. Everyone was really mad at me for shooting the bunny and we took it to the vet to
        get it fixed. The bill was only a few dollars. It was pretty cool that I could shoot a dart and hit something.

        This year I am going to get a bow and arrow set. Hopefully I can shoot that and not hurt anything. Maybe I will get
        it for Christmas.

    \end{displayquote}

    \begin{center}{\large{\textbf{Highly Prompted Computer Story: The Bunny}}}\end{center}
    \begin{displayquote}

        PROMPT:\textbf{One moment I regret in my life and will never forget} was a time I was visiting my
        grandmother's house for Christmas break. \textbf{I was young and naïve}. My cousin had just got this cool
        blow gun that could shoot darts from the other side. I took the gun outside to play with it for a bit.
        Out of the corner of my eye I saw a little bunny hop by and pointed the blow gun at the bunny.

        RESPONSE: I waited for the bunny to get close enough and shot it.

        The dart hit the bunny in the back and it fell over. I was shocked, scared and disgusted with myself. I ran
        over to the bunny, picked it up and the bunny was still alive. It was a fighter and I doubt that the dart
        did a lot of damage. I ran back inside and told my grandmother what happened and told her not to tell my
        parents. She agreed not to tell them. I was so ashamed of myself. I was ashamed that I did something like
        that and that I was never going to tell anyone. I felt as if I had betrayed my parents, my family and myself.

        I was so ashamed and I felt guilty. I was never going to do something like that again. I have never pointed
        a gun at anything again since that day. It was years before I told anyone what I did and I feel ashamed and
        guilty today.

    \end{displayquote}

    \hyperref[observations]{Return to Further Research}
        \begin{center}{\large{\textbf{Original Story: The Mouse}}}\end{center}
    \begin{displayquote}

    One time I was visiting my grandmother’s house in the mountains and we were getting ready to sit down for family dinner.
    Food was being pulled from the oven and the salad and cold foods were being placed on the table. As we were talking, a
    mouse raced across the carpet disturbing the peace in the house. I, being a little girl screamed because I was scared.
    My grandfather was determined to get rid of the mouse.

    He acted quick. As the mouse neared the dining table, my grandfather lifted his foot and slammed it down, killing the mouse.
    I was relieved, but the carpet was stained with blood. We cleaned the blood up and sat down for dinner and no one seemed
    to remember the mouse by the time dessert came.

    \end{displayquote}

        \begin{center}{\large{\textbf{Minimal Prompted Computer Story: The Mouse}}}\end{center}
    \begin{displayquote}

PROMPT: One time I was visiting my grandmother’s house in the mountains and we were getting ready to sit down for family
dinner. Food was being pulled from the oven and the salad and cold foods were being placed on the table. As we were
talking, a mouse raced across the carpet disturbing the peace in the house. I, being a little girl screamed because
I was scared. My grandfather was determined to get rid of the mouse.

RESPONSE: He went out to the garage to retrieve his .22. He returned to the kitchen and aimed the gun at the mouse.
He shot up in the air and said, “I don’t want to kill it, I just want it to leave.” He shot again, but this time he
hit the wall by the refrigerator and it was still alive. My grandfather then retrieved a broom, and began to hit the
mouse over and over. The mouse still was alive and not giving up. It jumped over my grandfather, and ran away.
My grandfather followed the mouse down the hall and out the front door. He made sure that the mouse was gone.

NOTE: \emph{It is interesting how this prompt caused the mouse to be viewed as the antagonist and my grandfather as the
protagonist. Even so, it frames humans as responding extremely violently in chaotic situations. Yet, it can also be
seen that human struggle and frustration is also painted in the response. The last line of “He made sure the mouse
was gone” shows that the AI understands that I as a little girl was scared and wanted relief. }

    \end{displayquote}

            \begin{center}{\large{\textbf{Highly Prompted Computer Story: The Mouse}}}\end{center}
    \begin{displayquote}

PROMPT: The first time I witnessed a death was when I was at my grandmother’s house for dinner. Some dishes were still
cooking in the oven, so we were sitting at the table talking. The conversation was calm and I felt safe with my family.
All of a sudden, a mouse came running across the floor. I was very scared and my grandfather was there to protect me
from the mouse. He wanted it gone.

RESPONSE: He reached down and grabbed the mouse by the tail. The mouse was very scared and he struggled to get away.
I started crying and my grandfather put the mouse down. I felt bad for the mouse and I was very scared. The next time
I witnessed death was when I was in the sixth grade. This was one of the hardest times in my life. It was the day of
my best friend’s father’s funeral. I was very sad about what happened to his father. I was sad that he had to go. I
was sad that he could not be there with his family. The next time I witnessed death was when I was at my grandmother’s
house and I heard a thud.

NOTE: \emph{Within this next prompt, I made sure to tell the same story with less atmosphere details (i.e., describing
dinner preparation) and instead focused more on emotion. I was attempting to show that family is supposed to make you
feel loved and safe and would go through any measures to protect you. A mouse would not be deemed as valuable as a human
life, and thus describing more how I felt during that moment was an attempt at provoking a more empathetic response. It
was interesting how in this response, the mouse was not painted clearly as the antagonist in the story. It focused more
on the aspect of death, which when considering empathy, is extremely appropriate (joint grieving).}

    \end{displayquote}

\begin{center}{\large{\textbf{Original Story: The Beach}}}\end{center}

\begin{displayquote}

I was once walking along an ocean beach with my dog. His name was Chester, and he was a big yellow labrador. I had
him on a short leash as we took a walk, but he often looked to explore more of the beach. He was pulling hard on the
leash and dragging me up into the dunes, away from the shoreline where the waves were breaking. As we were walking
back up into the sand dunes, I looked around and saw no one around. So I decided to take him off of his leash to
explore more of the beach. Suddenly, a woman with another dog came walking from over one of the hills from a direction
I neglected to look in. She had a dark black labrador that was also unleashed.

The other dog came running to where Chester and I stood. I was scared as my dog nervously growled at the black labrador.
Suddenly, as the other dog got closer, both started to look at each other with a look of familiarity and friendliness.
Then, suddenly, without warning, both went running off into the dunes. The woman and I looked at each other and tied to
figure out what to do next.

We both took off, running after the dogs, but found them right over the next dune. I asked the woman if her dog
often ran away, and she responded that he does not, as the house that they live in has a big white stone fence
that keeps him well contained. I knew of the fence she was referring to, and suddenly I realized that she was my
neighbor, and our dogs must know each other.

\end{displayquote}

\begin{center}{\large{\textbf{Minimal Prompted Computer Story: The Beach}}}\end{center}

\begin{displayquote}
PROMPT:I was once walking along an ocean beach with my dog. His name was Chester, and he was a big yellow labrador. I had
him on a short leash as we took a walk, but he often looked to explore more of the beach. He was pulling hard on the leash
and dragging me up into the dunes, away from the shoreline where the waves were breaking. As we were walking back up into
the sand dunes, I looked around and saw no one around. So I decided to take him off of his leash to explore more of the
beach. Suddenly, a woman with another dog came walking from over one of the hills from a direction I neglected to look in.
She had a dark black labrador that was also unleashed.

RESPONSE: The two dogs took off running towards each other. I heard the woman yell, "No! No! No!" as she was trying to
get her dog to stop. But it was too late. Her dog was already charging towards Chester, and I could see there was no
stopping them from colliding. They hit and started fighting. I ran over and grabbed Chester by his collar and pulled him
off of the other dog. The other dog got away. The woman was pulling on her dog and yelling at her dog, "Come on, come on!"
I asked her if she wanted me to call the police. She said that she did. The police came, and I gave them my information.
The woman did not get her dog's information. The police told me that they did not need it. But I took down the description
of her black lab, and the police did not ask me to give that information. They asked me if I wanted to press charges. I
told them that I did not want to. I did not want the police to put her in jail. I just wanted her to know that she was
not supposed to have her dog off of a leash and that people can get hurt. The police told me that she probably didn't
know and then said something about a leash law. I told them that I didn't know if there was a leash law in my city.

\end{displayquote}

\begin{center}{\large{\textbf{Highly Prompted Computer Story: The Beach}}}\end{center}

\begin{displayquote}

PROMPT: I was once walking along an ocean beach with my dog. His name was Chester, and he was a big yellow labrador. I had
him on a short leash as we took a walk, but he often looked to explore more of the beach. He was pulling hard on the
leash and dragging me up into the dunes, away from the shoreline where the waves were breaking. As we were walking
back up into the sand dunes, I looked around and saw no one around. So I decided to take him off of his leash to
explore more of the beach. Suddenly, a woman with another dog came walking from over one of the hills from a direction
I neglected to look in. She had a dark black labrador that was also unleashed.

The other dog came running to where Chester and I stood. I was scared as my dog nervously growled at the black labrador.
Suddenly, as the other dog got closer, both started to look at each other with a look of familiarity and friendliness.
Then, suddenly, without warning, both went running off into the dunes. The woman and I looked at each other and tied to
figure out what to do next.

RESPONSE: After a few minutes, we both decided to wait for the dogs in the dunes. They both ran off into a large sand
dune and disappeared over the top of it. We both sat waiting for them to come back over the top of the dune. After several
minutes, neither dog came over the top of the dune. The woman walked away and I waited alone for my dog to come back. A
few minutes later I heard a rustling over the top of the dune. I stood up and tried to see over the top of the dune. I
saw a yellow labrador running out of the dune and back down to the beach. I raised my hand and waved at the dog, thinking
it was Chester. Then, to my surprise, a black labrador came running out of the dune and down to the beach as well. Both d
ogs ran up to me and both of their owners came running to the dogs. Both owners were surprised to see their dogs with a
strange dog, and then light and laughter filled the air when both realized that their dogs had found each other on the
beach and had run off to play. The dogs stayed on the beach for about an hour, playing and swimming in the ocean. Both
owners were very thankful for the different outcome and both dogs were very excited to see each other again.

\end{displayquote}

    \section{\\Appendix B: Stages of Empathy}\label{appendix_b}

    \centerline{\includegraphics[width=0.8\textwidth]{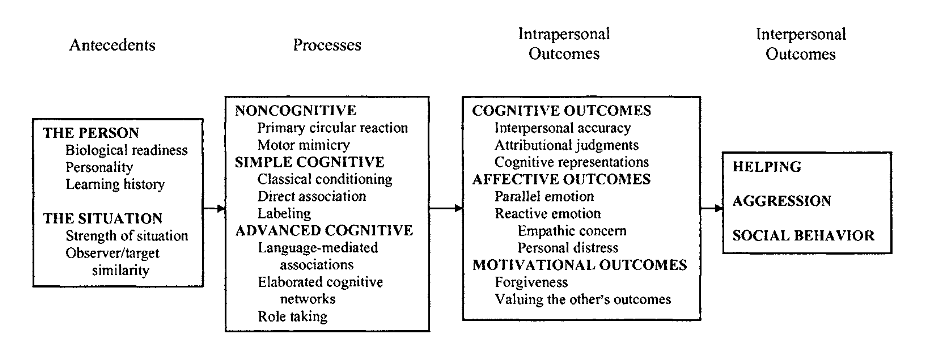}}
    \emph{\textbf{Empathy and its Components \cite{davis:empathy}}}
    \hyperref[definitions]{Return to Defining Empathy}
    \bibliography{main}{}
    \bibliographystyle{plain}

\end{document}